\def\year{2021}\relax

\documentclass[letterpaper]{article} 
\usepackage{aaai21}  
\usepackage{times}  
\usepackage{helvet} 
\usepackage{courier}  
\usepackage[hyphens]{url}  
\usepackage{graphicx} 
\usepackage{natbib}  
\usepackage{caption} 
\usepackage{microtype}
\usepackage{enumitem}
\usepackage{amsmath}
\usepackage{array}
\usepackage{url}            
\usepackage{microtype}
\usepackage{graphicx}
\usepackage{subfigure}
\usepackage{booktabs}
\usepackage{enumitem}
\usepackage{amsfonts}       
\usepackage{multirow}
\usepackage{array}
\usepackage{hhline}
\newcommand{\PreserveBackslash}[1]{\let\temp=\\#1\let\\=\temp}
\newcolumntype{C}[1]{>{\PreserveBackslash\centering}p{#1}}
\newcolumntype{R}[1]{>{\PreserveBackslash\raggedleft}p{#1}}
\newcolumntype{L}[1]{>{\PreserveBackslash\raggedright}p{#1}}
\newcolumntype{M}[1]{>{\centering\arraybackslash}m{#1}}

\title{TabNet: Attentive Interpretable Tabular Learning}

\author{%
 Sercan \"{O}. Ar{\i}k,
 Tomas Pfister \\ 
}

\affiliations{
 Google Cloud AI \\
 Sunnyvale, CA \\
 soarik@google.com, tpfister@google.com\\
}

\begin{document}
\maketitle

\begin{abstract}

We propose a novel high-performance and interpretable canonical deep tabular data learning architecture, TabNet. 
TabNet uses sequential attention to choose which features to reason from at each decision step, enabling interpretability and more efficient learning as the learning capacity is used for the most salient features.
We demonstrate that TabNet outperforms other variants on a wide range of non-performance-saturated tabular datasets and yields interpretable feature attributions plus insights into its global behavior. 
Finally, we demonstrate self-supervised learning for tabular data, significantly improving performance when unlabeled data is abundant.

\end{abstract}

\section{Introduction}
Deep neural networks (DNNs) have shown notable success with images \citep{resnet}, text \citep{rcnn_text} and audio \citep{deepspeech2}. 
For these, canonical architectures that efficiently encode the raw data into meaningful representations, fuel the rapid progress.
One data type that has yet to see such success with a canonical architecture is tabular data. 

\begin{figure*}[!h]
\centering
\includegraphics[width=0.99\textwidth]{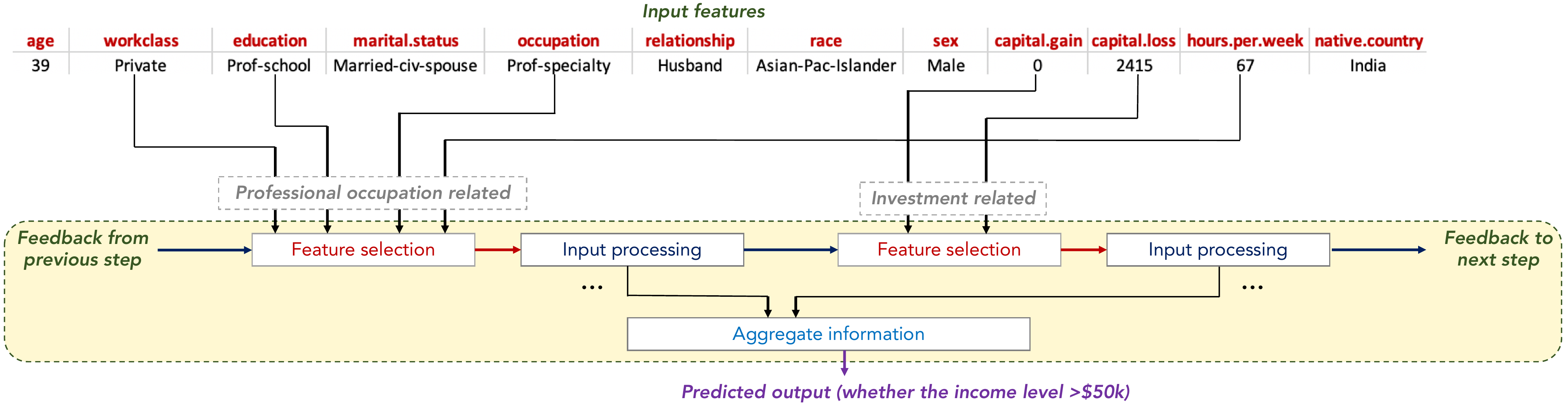}
\caption{TabNet's sparse feature selection exemplified for Adult Census Income prediction \citep{UCI}. 
Sparse feature selection enables interpretability and better learning as the capacity is used for the most salient features.
TabNet employs multiple decision blocks that focus on processing a subset of input features for reasoning. 
Two decision blocks shown as examples process features that are related to professional occupation and investments, respectively, in order to predict the income level. 
}
\label{fig:simplified_tabnet}
\end{figure*}

Despite being the most common data type in real-world AI (as it is comprised of any categorical and numerical features), \citep{mckinsey_report}, deep learning for tabular data remains under-explored, with variants of ensemble decision trees (DTs) still dominating most applications \citep{kaggletrends}.
Why?
First, because DT-based approaches have certain benefits: 
(i) they are representionally efficient for decision manifolds with approximately hyperplane boundaries which are common in tabular data; and
(ii) they are highly interpretable in their basic form (e.g. by tracking decision nodes) and there are popular post-hoc explainability methods for their ensemble form, e.g. \citep{shap} -- this is an important concern in many real-world applications;
(iii) they are fast to train. 
Second, because previously-proposed DNN architectures are not well-suited for tabular data: e.g. stacked convolutional layers or multi-layer perceptrons (MLPs) are vastly overparametrized -- the lack of appropriate inductive bias often causes them to fail to find optimal solutions for tabular decision manifolds \citep{Goodfellow2016, shavitt2018regularization, tablegan}. 

Why is deep learning worth exploring for tabular data? 
One obvious motivation is expected performance improvements particularly for large datasets \citep{deep_learning_scaling}. 
In addition, unlike tree learning,
DNNs enable gradient descent-based end-to-end learning for tabular data which can have a multitude of benefits: 
(i) efficiently encoding multiple data types like images along with tabular data;
(ii) alleviating the need for feature engineering, which is currently a key aspect in tree-based tabular data learning methods;
(iii) learning from streaming data
and perhaps most importantly
(iv) end-to-end models allow representation learning which enables many valuable application scenarios including data-efficient domain adaptation \citep{Goodfellow2016}, generative modeling \citep{unsupervised_rep_gan} and semi-supervised learning \citep{good_ssl}.

\begin{figure*}[!ht]

\centering
\includegraphics[width=0.9\textwidth]{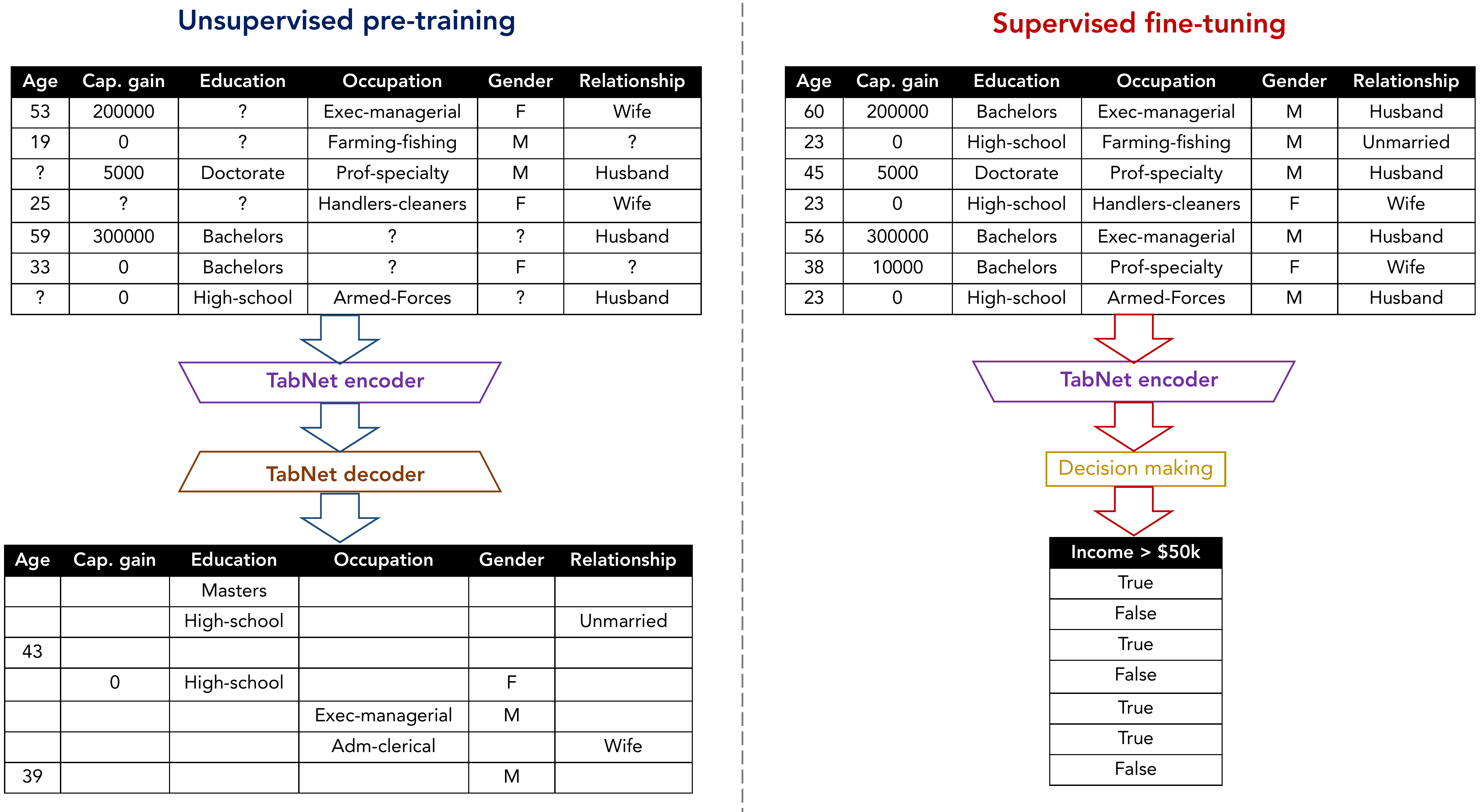}

\caption{Self-supervised tabular learning. Real-world tabular datasets have interdependent feature columns, e.g., the education level can be guessed from the occupation, or the gender can be guessed from the relationship. Unsupervised representation learning by masked self-supervised learning results in an improved encoder model for the supervised learning task.}
\label{fig:self_supervised}
\end{figure*}

We propose a new canonical DNN architecture for tabular data, TabNet.
The main contributions are summarized as:
\begin{enumerate}[noitemsep, nolistsep, leftmargin=*]
\item \emph{TabNet inputs raw tabular data without any preprocessing} and is \emph{trained using gradient descent-based optimization}, enabling flexible integration into end-to-end learning. 
\item \emph{TabNet uses sequential attention to choose which features to reason from at each decision step}, enabling interpretability and better learning as the learning capacity is used for the most salient features (see Fig. \ref{fig:simplified_tabnet}). 
This feature selection is \emph{instance-wise}, e.g. it can be different for each input, and unlike other instance-wise feature selection methods like \citep{l2x} or \citep{invase}, TabNet employs a \emph{single deep learning architecture for feature selection and reasoning}. 
\item Above design choices lead to two valuable properties: (i) \emph{TabNet outperforms or is on par with other tabular learning models} on various datasets for classification and regression problems from different domains; and (ii) \emph{TabNet enables two kinds of interpretability}: local interpretability that visualizes the importance of features and how they are combined, and global interpretability which quantifies the contribution of each feature to the trained model. 
\item Finally, \emph{for the first time for tabular data}, we show significant performance improvements by using unsupervised pre-training to predict masked features (see Fig. \ref{fig:self_supervised}). 

\end{enumerate}

\section{Related Work}

\noindent\newline{\bf Feature selection:} Feature selection broadly refers to judiciously picking a subset of features based on their usefulness for prediction. 
Commonly-used techniques such as forward selection and Lasso regularization \citep{feature_selection} attribute feature importance based on the entire training data, and are referred as \emph{global} methods. 
\emph{Instance-wise} feature selection refers to picking features individually for each input, studied in \citep{l2x} with an explainer model to maximize the mutual information between the selected features and the response variable, and in \citep{invase} by using an actor-critic framework to mimic a baseline while optimizing the selection. 
Unlike these, TabNet employs \emph{soft feature selection with controllable sparsity in end-to-end learning} -- a single model jointly performs feature selection and output mapping, resulting in superior performance with compact representations.
\begin{figure*}[!htbp]
\centering
\includegraphics[width=0.86\textwidth]{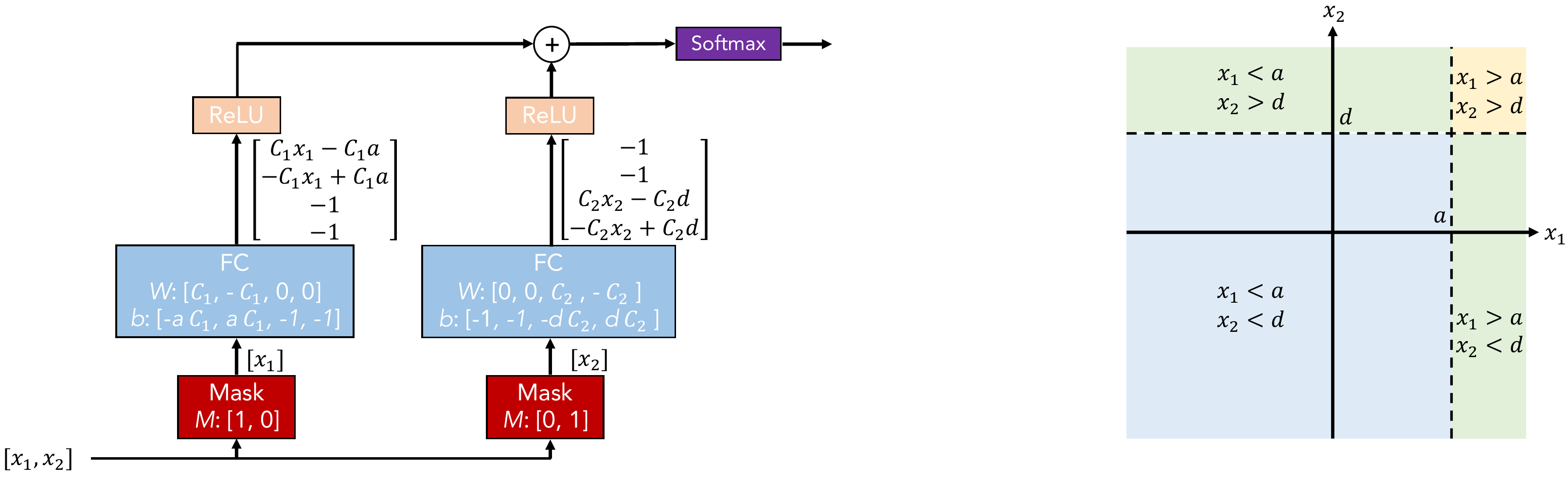}
\caption{
Illustration of DT-like classification using conventional DNN blocks (left) and the corresponding decision manifold (right).
Relevant features are selected by using multiplicative sparse masks on inputs. 
The selected features are linearly transformed, and after a bias addition (to represent boundaries) ReLU performs region selection by zeroing the regions. 
Aggregation of multiple regions is based on addition. 
As $C_1$ and $C_2$ get larger, the decision boundary gets sharper.}
\label{fig:dt_func}
\end{figure*}
\noindent\newline{\bf Tree-based learning:} DTs are commonly-used for tabular data learning. Their prominent strength is efficient picking of global features with the most statistical information gain \citep{decisiontree1}. 
To improve the performance of standard DTs, one common approach is ensembling to reduce variance. 
Among ensembling methods, random forests \citep{random_forest} use random subsets of data with randomly selected features to grow many trees. XGBoost \citep{XGBoost} and LightGBM \citep{lightgbm} are the two recent ensemble DT approaches that dominate most of the recent data science competitions. 
Our experimental results for various datasets show that tree-based models can be outperformed when the representation capacity is improved with deep learning while retaining their feature selecting property.
\noindent\newline{\bf Integration of DNNs into DTs:} 
Representing DTs with DNN building blocks as in \citep{DNN_init_tree} yields redundancy in representation and inefficient learning. 
Soft (neural) DTs \citep{neural_randomforest,neural_decisionforest} use differentiable decision functions, instead of non-differentiable axis-aligned splits. 
However, losing automatic feature selection often degrades performance. 
In \citep{dndt}, a soft binning function is proposed to simulate DTs in DNNs, by inefficiently enumerating of all possible decisions. 
\citep{tabnn} proposes a DNN architecture by explicitly leveraging expressive feature combinations, however, learning is based on transferring knowledge from gradient-boosted DT.
\citep{ant} proposes a DNN architecture by adaptively growing from primitive blocks while representation learning into edges, routing functions and leaf nodes. 
TabNet differs from these as it embeds soft feature selection with controllable sparsity via sequential attention. 
\noindent\newline{\bf Self-supervised learning:}
Unsupervised representation learning improves supervised learning especially in small data regime \citep{self_taught}. 
Recent work for text \citep{bert} and image \citep{selfie} data has shown significant advances -- driven by the judicious choice of the unsupervised learning objective (masked input prediction) and attention-based deep learning.

\section{TabNet for Tabular Learning}

DTs are successful for learning from real-world tabular datasets. 
With a specific design, conventional DNN building blocks can be used to implement DT-like output manifold, e.g. see Fig. \ref{fig:dt_func}).
In such a design, individual feature selection is key to obtain decision boundaries in hyperplane form, which can be generalized to a linear combination of features where coefficients determine the proportion of each feature. 
TabNet is based on such functionality and it outperforms DTs while reaping their benefits by careful design which:
(i) uses sparse instance-wise feature selection learned from data; 
(ii) constructs a sequential multi-step architecture, where each step contributes to a portion of the decision based on the selected features;
(iii) improves the learning capacity via non-linear processing of the selected features; and 
(iv) mimics ensembling via higher dimensions and more steps.

\begin{figure*}[!htbp]
\centering
\vspace{0cm}
\subfigure[TabNet encoder architecture]{\includegraphics[width=0.57\textwidth]{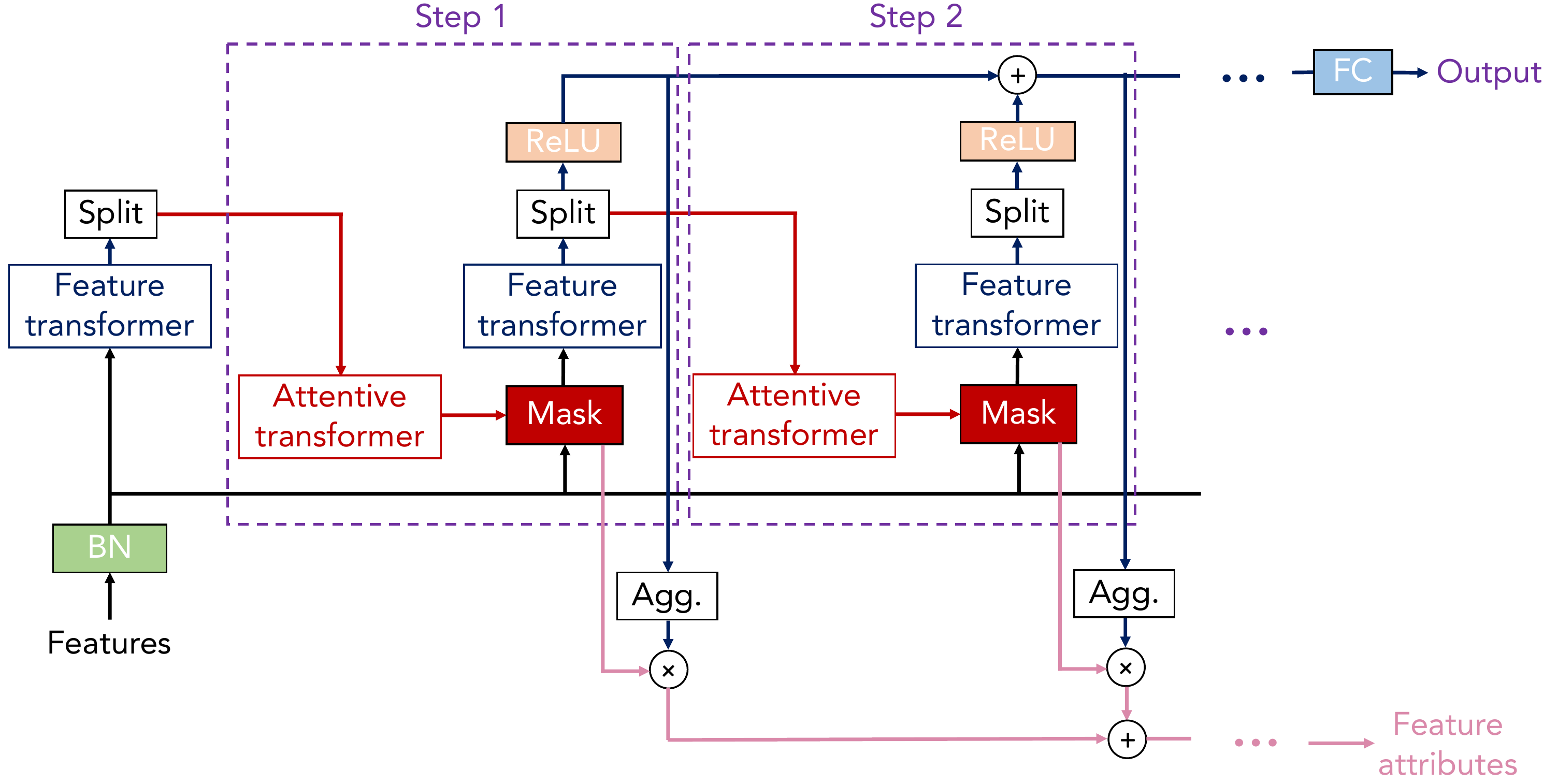}}
\vspace{0cm}
\subfigure[TabNet decoder architecture]{\includegraphics[width=0.4\textwidth]{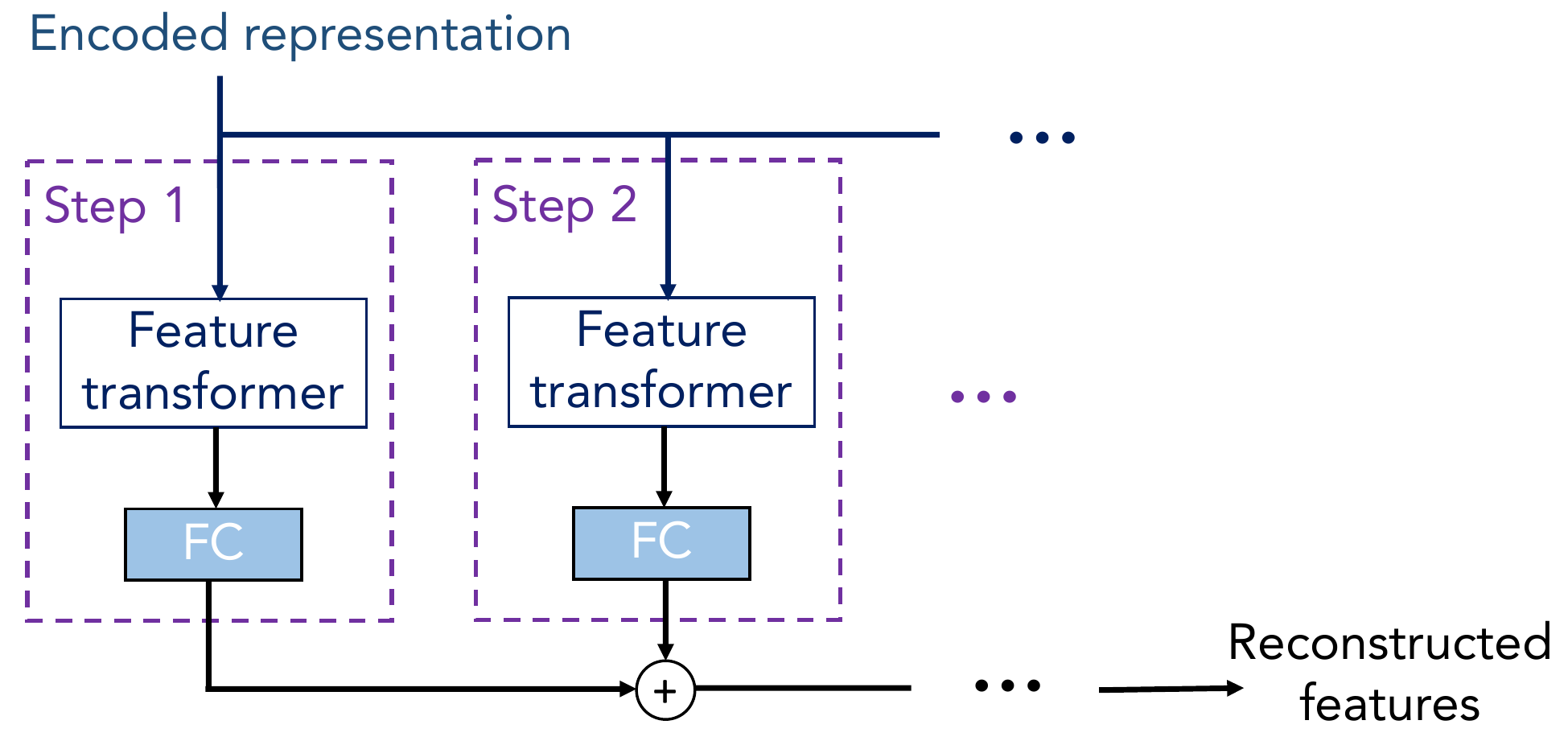}}
\vspace{0cm}
\subfigure[ ]{\includegraphics[width=0.6\textwidth]{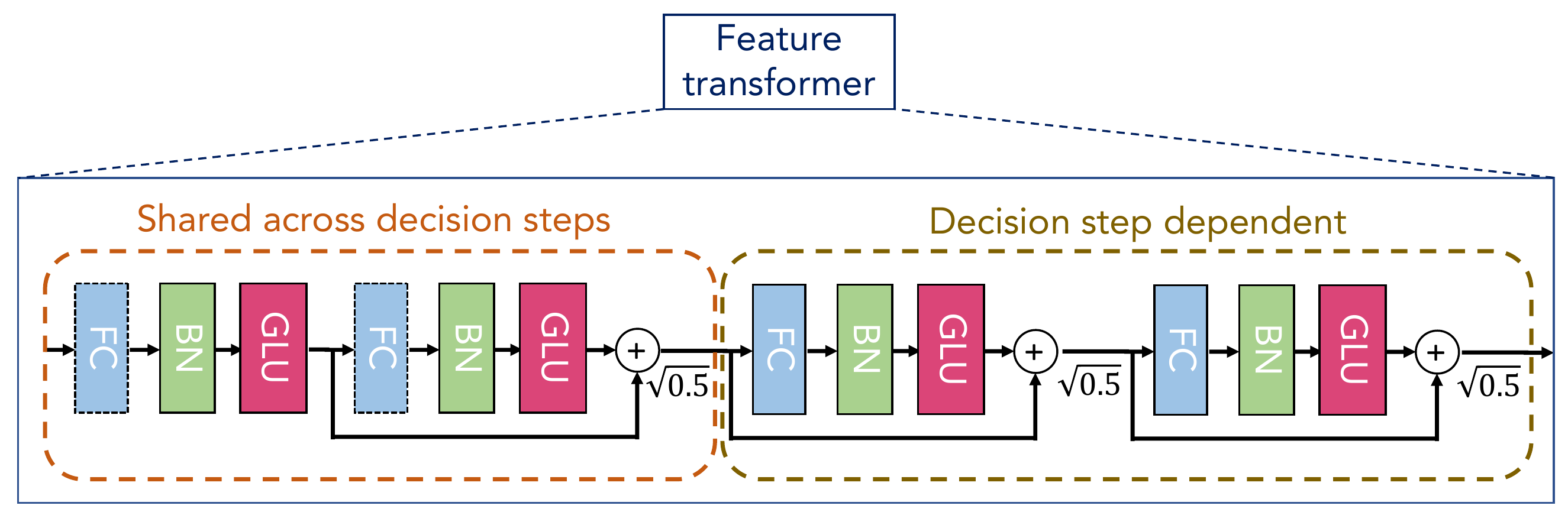}}
\hspace{0.3cm}
\subfigure[ ]{\includegraphics[width=0.2\textwidth]{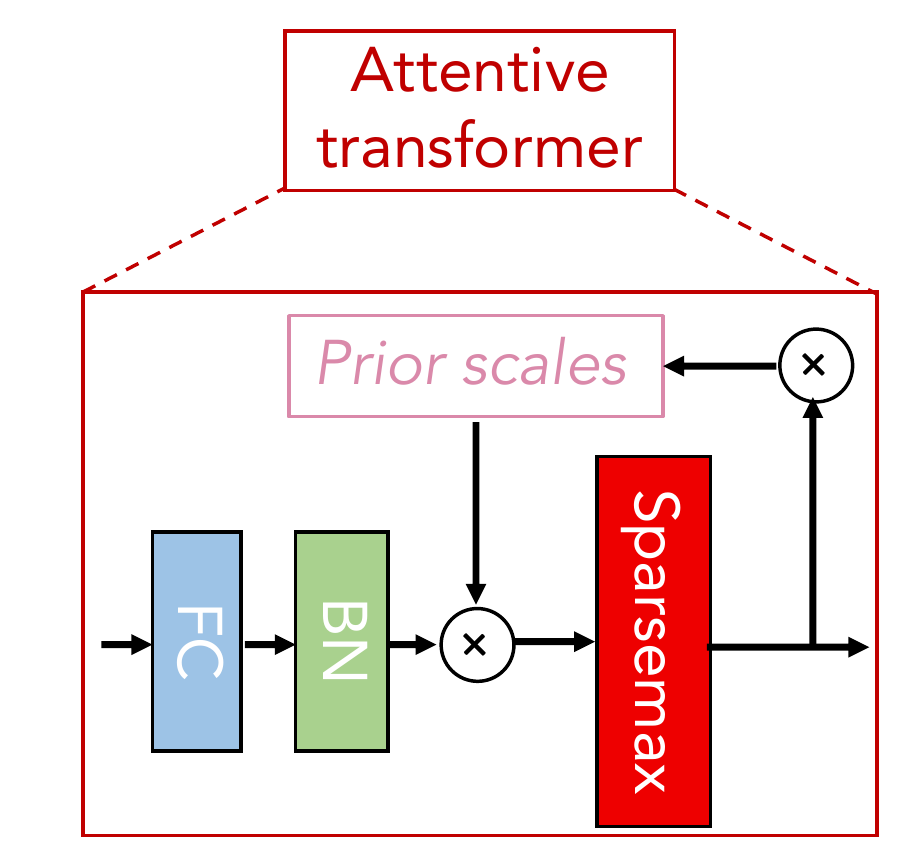}}
\caption{(a) TabNet encoder, composed of a feature transformer, an attentive transformer and feature masking. 
A split block divides the processed representation to be used by the attentive transformer of the subsequent step as well as for the overall output. 
For each step, the feature selection mask provides interpretable information about the model's functionality, and the masks can be aggregated to obtain global feature important attribution. 
(b) TabNet decoder, composed of a feature transformer block at each step. 
(c) A feature transformer block example -- 4-layer network is shown, where 2 are shared across all decision steps and 2 are decision step-dependent. 
Each layer is composed of a fully-connected (FC) layer, BN and GLU nonlinearity. 
(d) An attentive transformer block example -- a single layer mapping is modulated with a prior scale information which aggregates how much each feature has been used before the current decision step. 
sparsemax \citep{sparsemax} is used for normalization of the coefficients, resulting in sparse selection of the salient features.
\vspace{0cm}
}

\label{fig:tabnet}
\label{fig:feature_transform}
\label{fig:attn_transform}
\end{figure*}

Fig. \ref{fig:tabnet} shows the TabNet architecture for encoding tabular data. 
We use the raw numerical features and consider mapping of categorical features with trainable embeddings.
We do not consider any global feature normalization, but merely apply batch normalization (BN). 
We pass the same $D$-dimensional features $\mathbf{f} \in \Re ^ {B \times D}$ to each decision step, where $B$ is the batch size. 
TabNet's encoding is based on sequential multi-step processing with $N_{steps}$ decision steps. 
The $i^{th}$ step inputs the processed information from the $(i-1)^{th}$ step to decide which features to use and outputs the processed feature representation to be aggregated into the overall decision. 
The idea of top-down attention in the sequential form is inspired by its applications in processing visual and text data \citep{compositional_attention} and reinforcement learning \citep{S3A} while searching for a small subset of relevant information in high dimensional input. 
\noindent\newline{\bf Feature selection:}
We employ a learnable mask $\mathbf{M[i]} \in \Re ^ {B \times D}$ for soft selection of the salient features. Through sparse selection of the most salient features, the learning capacity of a decision step is not wasted on irrelevant ones, and thus the model becomes more parameter efficient. The masking is multiplicative, $\mathbf{M[i]} \cdot \mathbf{f}$. We use an attentive transformer (see Fig. \ref{fig:attn_transform}) to obtain the masks using the processed features from the preceding step, $\mathbf{a[i-1]}$:
$
\mathbf{M[i]} = \text{sparsemax}(\mathbf{P[i-1]} \cdot \text{h}_i(\mathbf{a[i-1]})).
$
Sparsemax normalization \citep{sparsemax} encourages sparsity by mapping the Euclidean projection onto the probabilistic simplex, which is observed to be superior in performance and aligned with the goal of sparse feature selection for explainability. Note that $\sum\nolimits_{j=1}^{D} \mathbf{M[i]_{b,j}} = 1$. $\text{h}_i$ is a trainable function, shown in Fig. \ref{fig:attn_transform} using a FC layer, followed by BN. $\mathbf{P[i]}$ is the prior scale term, denoting how much a particular feature has been used previously:
$
\mathbf{P[i]} = \prod\nolimits_{j=1}^{i} (\gamma - \mathbf{M[j]}), 
$
where $\gamma$ is a relaxation parameter -- when $\gamma=1$, a feature is enforced to be used only at one decision step and as $\gamma$ increases, more flexibility is provided to use a feature at multiple decision steps. $\mathbf{P[0]}$ is initialized as all ones, $\mathbf{1} ^ {B \times D}$, without any prior on the masked features. If some features are unused (as in self-supervised learning), corresponding $\mathbf{P[0]}$ entries are made 0 to help model's learning.
To further control the sparsity of the selected features, we propose sparsity regularization in the form of entropy  \citep{entropy_ssl},
$L_{sparse} = \sum\nolimits_{i=1}^{N_{steps}} \sum\nolimits_{b=1}^{B} \sum\nolimits_{j=1}^{D} \frac{-\mathbf{M_{b,j}[i]} \log(\mathbf{M_{b,j}[i]} \! +\!  \epsilon)}{N_{steps} \cdot B},$
where $\epsilon$ is a small number for numerical stability. We add the sparsity regularization to the overall loss, with a coefficient $\lambda_{sparse}$. Sparsity provides a favorable inductive bias for datasets where most features are redundant.
\noindent\newline{\bf Feature processing:}
We process the filtered features using a feature transformer (see Fig. \ref{fig:feature_transform}) and then split for the decision step output and information for the subsequent step,
$[\mathbf{d[i]}, \mathbf{a[i]}] = \text{f}_i(\mathbf{M[i]} \cdot \mathbf{f})$,
where $\mathbf{d[i]} \in \Re ^ {B \times N_d}$ and $\mathbf{a[i]} \in \Re ^ {B \times N_a}$. For parameter-efficient and robust learning with high capacity, a feature transformer should comprise layers that are shared across all decision steps (as the same features are input across different decision steps), as well as decision step-dependent layers. Fig. \ref{fig:attn_transform} shows the implementation as concatenation of two shared layers and two decision step-dependent layers. Each FC layer is followed by BN and gated linear unit (GLU) nonlinearity \citep{glu},
eventually connected to a normalized residual connection with normalization. Normalization with $\sqrt{0.5}$ helps to stabilize learning by ensuring that the variance throughout the network does not change dramatically \citep{convseq2seq}. For faster training, we use large batch sizes with BN. Thus, except the one applied to the input features, we use ghost BN \citep{ghost_batch_norm} form, using a virtual batch size $B_V$ and momentum $m_B$. For the input features, we observe the benefit of low-variance averaging and hence avoid ghost BN. 
Finally, inspired by decision-tree like aggregation as in Fig. \ref{fig:dt_func}, we construct the overall decision embedding as $\mathbf{d_{out}} = \sum\nolimits_{i=1}^{N_{steps}} \text{ReLU}(\mathbf{d[i]})$.
We apply a linear mapping $\mathbf{W_{final}} \mathbf{d_{out}}$ to get the output mapping.\footnote{For discrete outputs, we additionally employ softmax during training (and argmax during inference).}
\noindent\newline{\textbf{Interpretability:}}
TabNet's feature selection masks can shed light on the selected features at each step. 
If $\mathbf{M_{b,j}[i]} = 0$, then $j^{th}$ feature of the $b^{th}$ sample should have no contribution to the decision. 
If $\text{f}_i$ were a linear function, the coefficient $\mathbf{M_{b,j}[i]}$ would correspond to the feature importance of $\mathbf{f_{b,j}}$. 
Although each decision step employs non-linear processing, their outputs are combined later in a linear way. 
We aim to quantify an aggregate feature importance in addition to analysis of each step. 
Combining the masks at different steps requires a coefficient that can weigh the relative importance of each step in the decision. 
We simply propose $\mathbf{\eta_b[i]} =\sum_{c=1}^{N_d} \text{ReLU}(\mathbf{d_{b,c}[i]})$ to denote the aggregate decision contribution at $i^{th}$ decision step for the $b^{th}$ sample.
Intuitively, if $\mathbf{d_{b,c}[i]} < 0$, then all features at $i^{th}$ decision step should have 0 contribution to the overall decision.
As its value increases, it plays a higher role in the overall linear combination. Scaling the decision mask at each decision step with $\mathbf{\eta_b[i]}$, we propose the aggregate feature importance mask,
$\mathbf{M_{agg-b,j}} \! = \! \sum\nolimits_{i=1}^{N_{steps}}\! \mathbf{\eta_{b}[i]} \mathbf{M_{b,j}[i]} \Big/ \sum\nolimits_{j=1}^{D} \! \sum\nolimits_{i=1}^{N_{steps}} \! \mathbf{\eta_{b}[i]} \mathbf{M_{b,j}[i]}.$\footnote{Normalization is used to ensure $\sum\nolimits_{j=1}^{D} \mathbf{M_{agg-b,j}}=1$.}

\begin{table*}[!htbp]
\caption{Mean and std. of test area under the receiving operating characteristic curve (AUC) on 6 synthetic datasets from \citep{l2x}, for TabNet vs. other feature selection-based DNN models: No sel.: using all features without any feature selection, Global: using only globally-salient features, Tree Ensembles \citep{Geurts2006}, Lasso-regularized model, L2X \citep{l2x} and INVASE \citep{invase}. Bold numbers denote the best for each dataset.}
\centering
\begin{tabular}{|l|l|l|l|l|l|l|}
\hline
\multirow{2}{*}{\textit{Model}}& \multicolumn{6}{|c|}{\textit{Test AUC}} \\  \cline{2-7}
& \multicolumn{1}{c|}{Syn1} & \multicolumn{1}{c|}{Syn2} & \multicolumn{1}{c|}{Syn3} & \multicolumn{1}{c|}{Syn4} & \multicolumn{1}{c|}{Syn5} & \multicolumn{1}{c|}{Syn6} \\ \hline
No selection & .578 $\pm$ .004 & .789 $\pm$ .003 & .854 $\pm$ .004 & .558 $\pm$ .021 & .662 $\pm$ .013 & .692 $\pm$ .015 \\ \hline
Tree & .574 $\pm$ .101 & .872 $\pm$ .003 & .899 $\pm$ .001 & .684 $\pm$ .017 & .741 $\pm$ .004 & .771 $\pm$ .031 \\ \hline
Lasso-regularized & .498 $\pm$ .006 & .555 $\pm$ .061 & .886 $\pm$ .003 & .512 $\pm$ .031 & .691 $\pm$ .024 & .727 $\pm$ .025 \\ \hline
L2X & .498 $\pm$ .005 & .823 $\pm$ .029 & .862 $\pm$ .009 & .678 $\pm$ .024 & .709 $\pm$ .008 & .827 $\pm$ .017 \\ \hline
INVASE & \textbf{.690 $\pm$ .006} & .877 $\pm$ .003 & \textbf{.902 $\pm$ .003} & \textbf{.787 $\pm$ .004} & .784 $\pm$ .005 & .877 $\pm$ .003 \\ \hline
Global & .686 $\pm$ .005 & .873 $\pm$ .003 & .900 $\pm$ .003 & .774 $\pm$ .006 & .784 $\pm$ .005 & .858 $\pm$ .004 \\ \hline
\textit{TabNet} & .682 $\pm$ .005 & \textbf{.892 $\pm$ .004} & .897 $\pm$ .003 & .776 $\pm$ .017 & \textbf{.789 $\pm$ .009} & \textbf{.878 $\pm$ .004} \\ \hline
\end{tabular}
\vspace{0cm}
\label{table:feature_selection}
\end{table*}
\vspace{0cm}
\noindent\newline{\textbf{Tabular self-supervised learning:}}
We propose a decoder architecture to reconstruct tabular features from the TabNet encoded representations. The decoder is composed of feature transformers, followed by FC layers at each decision step. The outputs are summed to obtain the reconstructed features.
We propose the task of prediction of missing feature columns from the others. Consider a binary mask $\mathbf{S} \in \{ 0, 1 \} ^ {B \times D}$.  
The TabNet encoder inputs $(\mathbf{1} - \mathbf{S}) \cdot \mathbf{\hat{f}}$ and the decoder outputs the reconstructed features, $\mathbf{S} \cdot \mathbf{\hat{f}}$. 
We initialize $\mathbf{P[0]} = (\mathbf{1} - \mathbf{S})$ in encoder so that the model emphasizes merely on the known features, and the decoder's last FC layer is multiplied with $\mathbf{S}$ to output the unknown features. 
We consider the reconstruction loss in self-supervised phase:
$\sum_{b=1}^{B} \sum_{j=1}^{D} \left | 
\frac{ ({\mathbf{\hat{f}_{b,j}}} - \mathbf{f_{b,j}}) \cdot \mathbf{\mathbf{S_{b,j}}}}
{\sqrt{\sum_{b=1}^{B} (\mathbf{f_{b,j}} - {1/B} \sum_{b=1}^{B} \mathbf{f_{b,j}})^{2}}}
\right | ^2.
$
Normalization with the population standard deviation of the ground truth is beneficial, as the features may have different ranges. We sample $\mathbf{S_{b,j}}$ independently from a Bernoulli distribution with parameter $p_s$, at each iteration. 

\section{Experiments}
\vspace{0cm}

We study TabNet in wide range of problems, that contain regression or classification tasks, \emph{particularly with published benchmarks}. 
For all datasets, categorical inputs are mapped to a single-dimensional trainable scalar with a learnable embedding\footnote{In some cases, higher dimensional embeddings may slightly improve the performance, but interpretation of individual dimensions may become challenging.} and numerical columns are input without and preprocessing.\footnote{Specially-designed feature engineering, e.g. logarithmic transformation of variables highly-skewed distributions, may further improve the results but we leave it out of the scope of this paper.}
We use standard classification (softmax cross entropy) and regression (mean squared error) loss functions and we train until convergence. Hyperparameters of the TabNet models are optimized on a validation set and listed in Appendix. TabNet performance is not very sensitive to most hyperparameters as shown with ablation studies in Appendix. In Appendix, we also present ablation studies on various design and guidelines on selection of the key hyperparameters.
For all experiments we cite, we use the same training, validation and testing data split with the original work. Adam optimization algorithm \citep{adam} and Glorot uniform initialization are used for training of all models.\footnote{An open-source implementation will be released.}

\subsection{Instance-wise feature selection}
\label{performance_section}
\vspace{0cm}
Selection of the salient features is crucial for high performance, especially for small datasets. We consider 6 tabular datasets from \citep{l2x} (consisting 10k training samples). The datasets are constructed in such a way that only a subset of the features determine the output. For Syn1-Syn3, salient features are same for all instances (e.g., the output of Syn2 depends on features $X_3$-$X_6$), and global feature selection, as if the salient features were known, would give high performance. For Syn4-Syn6, salient features are instance dependent (e.g., for Syn4, the output depends on either $X_1$-$X_2$ or $X_3$-$X_6$ depending on the value of $X_{11}$), which makes global feature selection suboptimal.
Table \ref{table:feature_selection} shows that TabNet outperforms others (Tree Ensembles \citep{Geurts2006}, LASSO regularization, L2X \citep{l2x}) and is on par with INVASE \citep{invase}. For Syn1-Syn3, TabNet performance is close to global feature selection - \emph{it can figure out what features are globally important}. For Syn4-Syn6, eliminating instance-wise redundant features, TabNet improves global feature selection.
All other methods utilize a predictive model with 43k parameters, and the total number of parameters is 101k for INVASE due to the two other models in the actor-critic framework. TabNet is a single architecture, and its size is 26k for Syn1-Syn3 and 31k for Syn4-Syn6. The compact representation is one of TabNet's valuable properties. 

\vspace{0cm}
\subsection{Performance on real-world datasets} 

\bgroup
\def\arraystretch{0.95}
\begin{table}[h!]
\vspace{0cm}
\caption{Performance for Forest Cover Type dataset.}
\vspace{0cm}
\centering
\begin{tabular}{|C{4.9 cm}|C{2.5 cm}|}
\cline{1-2}
\textit{Model} & \textit{Test accuracy (\%)}        \\ \cline{1-2}
XGBoost &  89.34                \\ \cline{1-2}
LightGBM  & 89.28              \\ \cline{1-2}
CatBoost & 85.14              \\ \cline{1-2} 
AutoML Tables  & 94.95                  \\ \cline{1-2}
\textit{TabNet}  &  \textbf{96.99}  \\ \cline{1-2}
\end{tabular}
\label{table:covertype}
\end{table}
\egroup

\noindent\newline{\bf Forest Cover Type \citep{UCI}:} The task is classification of forest cover type from cartographic variables. Table \ref{table:covertype} shows that TabNet outperforms ensemble tree based approaches that are known to achieve solid performance \citep{xgboost_gpu}. 
We also consider AutoML Tables \citep{automl_tables}, an automated search framework based on ensemble of models including DNN, gradient boosted DT, AdaNet \citep{adanet} and ensembles \citep{automl_tables} with very thorough hyperparameter search. 
A single TabNet without fine-grained hyperparameter search outperforms it.

\begin{table}[h!]
\vspace{0cm}
\caption{Performance for Poker Hand induction dataset.}
\vspace{0cm}
\centering
\begin{tabular}{|C{4.0 cm}|C{2.8 cm}|}
\cline{1-2}
    \textit{Model} & \textit{Test accuracy (\%)}        \\ \cline{1-2}
    DT &  50.0                \\ \cline{1-2}
    MLP & 50.0               \\ \cline{1-2}
    Deep neural DT  & 65.1   \\
    \cline{1-2}
    XGBoost &  71.1                \\ \cline{1-2} 
    LightGBM &  70.0                \\ \cline{1-2} 
    CatBoost &  66.6                \\ \hhline{|=|=|}
    \textit{TabNet}  & \textbf{99.2}  \\ \hhline{|=|=|}
    Rule-based  &  100.0                 \\ \cline{1-2}
\end{tabular}
\label{table:poker}
\vspace{0cm}
\end{table}

\vspace{0cm}
\noindent\newline{\bf Poker Hand \citep{UCI}:} The task is classification of the poker hand from the raw suit and rank attributes of the cards. The input-output relationship is deterministic and hand-crafted rules can get 100\% accuracy. Yet, conventional DNNs, DTs, and even their hybrid variant of deep neural DTs \cite{dndt} severely suffer from the imbalanced data and cannot learn the required sorting and ranking operations \cite{dndt}. Tuned XGBoost, CatBoost, and LightGBM show very slight improvements over them. TabNet outperforms other methods, as it can perform highly-nonlinear processing with its depth, without overfitting thanks to instance-wise feature selection.

\begin{table}[h!]
\vspace{0cm}
\caption{Performance on Sarcos dataset. Three TabNet models of different sizes are considered. }
\vspace{0cm}
\centering
\begin{tabular}{|C{3.7 cm}|C{1.5 cm}|C{1.7 cm}|}
\cline{1-3}
    \textit{Model} & \textit{Test MSE}  & \textit{Model size}        \\ \cline{1-3}
    Random forest & 2.39 &   16.7K  \\ \cline{1-3}
    Stochastic DT & 2.11  &   28K     \\ \cline{1-3}
    MLP & 2.13 &   0.14M     \\ \cline{1-3}
    Adaptive neural tree  &  1.23  &  0.60M  \\ \cline{1-3}
    Gradient boosted tree & 1.44   &   0.99M     \\ \hhline{|=|=|=|}
    \textit{TabNet-S}  &  \textbf{1.25} & \textbf{6.3K} \\ \cline{1-3}
    \textit{TabNet-M}  &  \textbf{0.28} & \textbf{0.59M} \\ \cline{1-3}
    \textit{TabNet-L}  &  \textbf{0.14} & \textbf{1.75M} \\ \cline{1-3}
\end{tabular}
\label{table:sarcos}
\vspace{0cm}
\end{table}

\vspace{0cm}
\noindent\newline{\bf Sarcos \citep{sarcos_dataset}:} The task is regressing inverse dynamics of an anthropomorphic robot arm. 
\citep{ant} shows that decent performance with a very small model is possible with a random forest. 
In the very small model size regime, TabNet's performance is on par with the best model from \citep{ant} with ~100x more parameters. 
When the model size is not constrained, TabNet achieves almost an order of magnitude lower test MSE. 

\begin{table}[h!]
\vspace{0cm}
\caption{Performance on Higgs Boson dataset. Two TabNet models are denoted with -S and -M.}
\vspace{0cm}
\centering
\begin{tabular}{|C{3.6 cm}|C{1.8 cm}|C{1.6 cm}|}
\cline{1-3}
    \textit{Model} & \textit{Test acc.} (\%)  & \textit{Model size}       \\ \cline{1-3}
    Sparse evolutionary MLP  & \textbf{78.47}  &   \textbf{81K}     \\ \cline{1-3}
    Gradient boosted tree-S  & 74.22   &  0.12M      \\ \cline{1-3}
    Gradient boosted tree-M  & 75.97   &     0.69M   \\ \cline{1-3}
    MLP   &  78.44 &  2.04M  \\ \cline{1-3}
    Gradient boosted tree-L  &  76.98  &    6.96M    \\ \hhline{|=|=|=|}
    \textit{TabNet-S}  &  78.25 & 81K \\ \cline{1-3}
    \textit{TabNet-M}  &  \textbf{78.84} & \textbf{0.66M} \\ \cline{1-3}
\end{tabular}
\label{table:higgs}
\vspace{0cm}
\end{table}

\vspace{0cm}
\noindent\newline{\bf Higgs Boson \citep{UCI}:} The task is distinguishing between a Higgs bosons process vs. background. 
Due to its much larger size (10.5M instances), DNNs outperform DT variants even with very large ensembles. 
TabNet outperforms MLPs with more compact representations. 
We also compare to the state-of-the-art evolutionary sparsification algorithm \citep{sparse_mlp} that integrates non-structured sparsity into training.
With its compact representation, TabNet yields almost similar performance to sparse evolutionary training for the same number of parameters.
Unlike sparse evolutionary training, the sparsity of TabNet is structured -- it does not degrade the operational intensity \citep{structured_sparsity} and can efficiently utilize modern multi-core processors.

\begin{table}[h!]
\vspace{0cm}
\caption{Performance for Rossmann Store Sales dataset.}
\centering
\begin{tabular}{|C{3 cm}|C{2.7 cm}|}
\cline{1-2}
    \textit{Model} & \textit{Test MSE}        \\ \cline{1-2}
    MLP & 512.62              \\ \cline{1-2}
    XGBoost & 490.83                 \\ \cline{1-2}
    LightGBM  & 504.76                \\ \cline{1-2}
    CatBoost & 489.75               \\ \hhline{|=|=|}
    \textit{TabNet}  &  \textbf{485.12}  \\ \cline{1-2}
\end{tabular}
\label{table:rossmann}
\vspace{0cm}
\end{table}
\vspace{0cm}

\noindent\newline{\bf Rossmann Store Sales \citep{kaggle_rossmann}:} The task is forecasting the store sales from static and time-varying features. We observe that TabNet outperforms commonly-used methods. 
The time features (e.g. day) obtain high importance, and the benefit of instance-wise feature selection is observed for cases like holidays where the sales dynamics are different.

\vspace{0cm}
\subsection{Interpretability}
\label{interpretability_section}
\vspace{0cm}

\begin{figure*}[!htbp]
\centering
\includegraphics[width=0.55\textwidth]{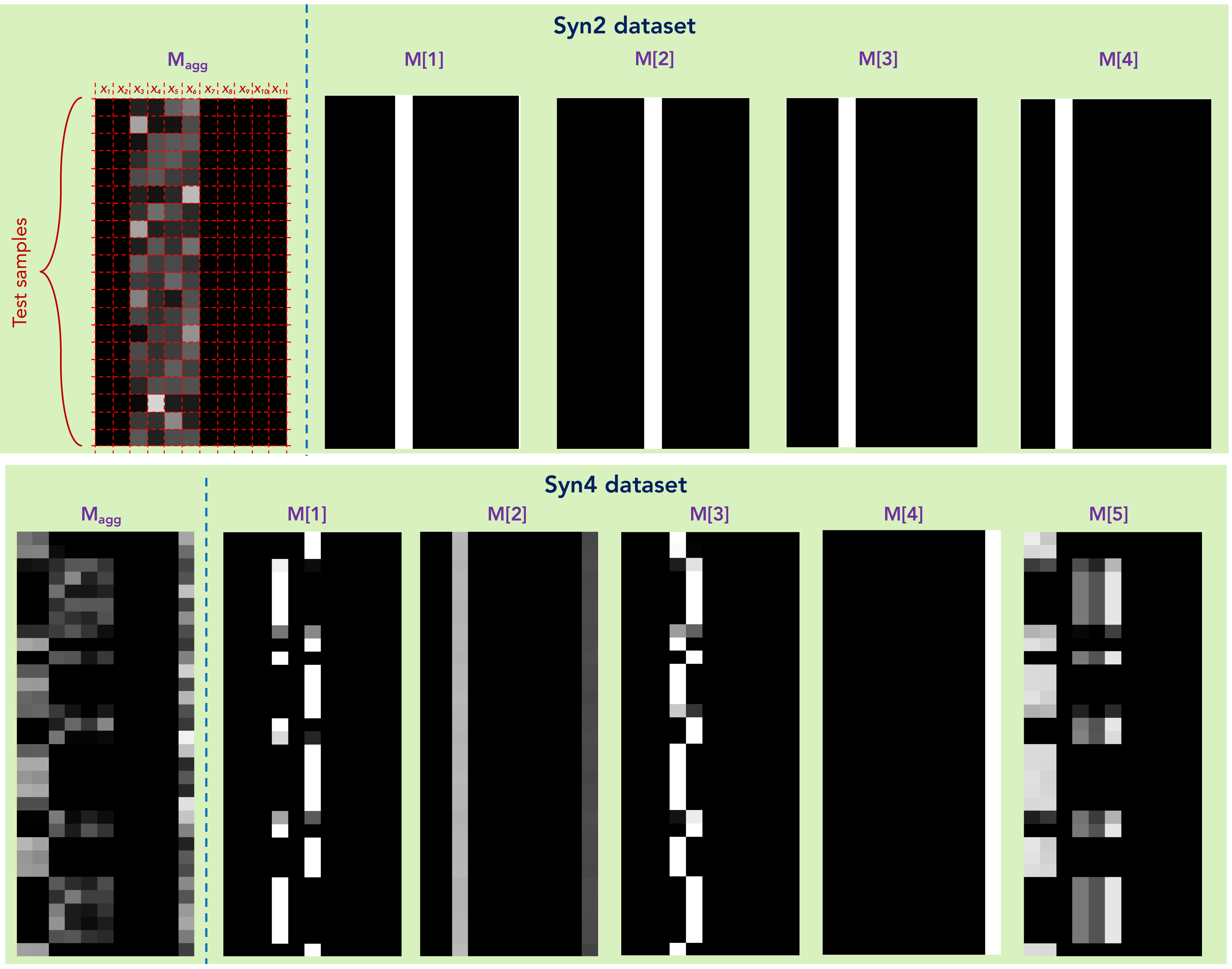}

\caption{Feature importance masks $\mathbf{M[i]}$ (that indicate feature selection at $i^{th}$ step) and the aggregate feature importance mask $\mathbf{M_{agg}}$ showing the global instance-wise feature selection, on Syn2 and Syn4 \citep{l2x}. Brighter colors show a higher value. E.g. for Syn2, only $X_3$-$X_6$ are used.}.

\label{fig:syn_masks}
\end{figure*}

\vspace{0cm}
\noindent\newline{\bf Synthetic datasets:} Fig. \ref{fig:syn_masks} shows the aggregate feature importance masks for the synthetic datasets from Table \ref{table:feature_selection}.\footnote{For better illustration here, the models are trained with 10M samples rather than 10K as we obtain sharper selection masks.} The output on Syn2 only depends on $X_3$-$X_6$ and we observe that the aggregate masks are almost all zero for irrelevant features and TabNet merely focuses on the relevant ones. 
For Syn4, the output depends on either $X_1$-$X_2$ or $X_3$-$X_6$ depending on the value of $X_{11}$. 
TabNet yields accurate instance-wise feature selection -- it allocates a mask to focus on the indicator $X_{11}$, and assigns almost all-zero weights to irrelevant features (the ones other than two feature groups).

\begin{figure}[!htbp]
\centering
\includegraphics[width=0.45\textwidth]{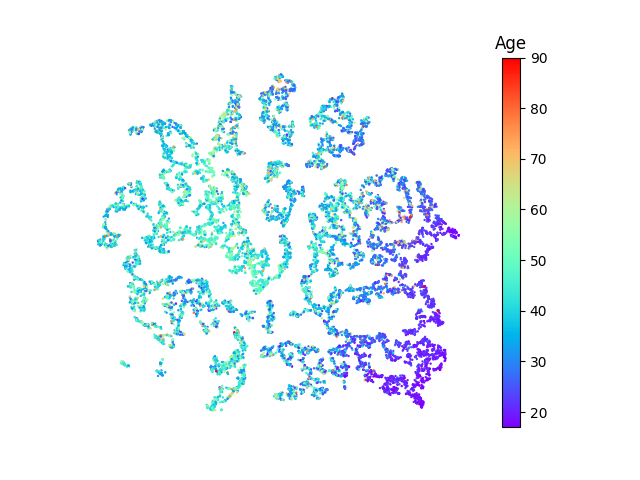}
\caption{First two dimensions of the T-SNE of the decision manifold for Adult and the impact of the top feature `Age'.}
\label{fig:tsne}
\end{figure}

\vspace{0cm}
\noindent\newline{\bf Real-world datasets:} We first consider the simple task of mushroom edibility prediction \citep{UCI}. TabNet achieves 100\% test accuracy on this dataset. It is indeed known \citep{UCI} that ``Odor" is the most discriminative feature -- with ``Odor" only, a model can get $>98.5 \%$ test accuracy \citep{UCI}. Thus, a high feature importance is expected for it. TabNet assigns an importance score ratio of 43\% for it, while other methods like LIME \citep{lime}, Integrated Gradients \citep{integrated_gradients} and DeepLift \citep{deeplift} assign less than 30\% \citep{global_explanations}.
Next, we consider Adult Census Income.
TabNet yields feature importance rankings consistent with the well-known \citep{shap, nbviewer} (see Appendix) 
For the same problem, Fig. \ref{fig:tsne} shows the clear separation between age groups, as suggested by ``Age" being the most important feature by TabNet. 


\begin{figure}[!htbp]
\centering
\includegraphics[width=0.45\textwidth]{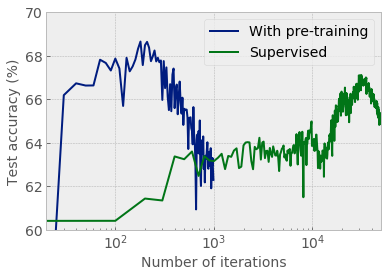}
\vspace{0cm}
\caption{Training curves on Higgs dataset with 10k samples.}
\label{fig:ssl_training}
\end{figure}

\subsection{Self-supervised learning}
\vspace{0cm}

\begin{table}[!htbp]
\caption{Mean and std. of accuracy (over 15 runs) on Higgs with Tabnet-M model, varying the size of the training dataset for supervised fine-tuning.}
\vspace{0cm}
\centering
\begin{tabular}{|C{1.8 cm}|C{2.5 cm}|C{2.5 cm}|}
\cline{1-3}
\textit{Training} & \multicolumn{2}{|c|}{\textit{Test accuracy} (\%)}       \\ \cline{2-3}
\textit{dataset size}  & \textit{Supervised} & \textit{With pre-training} \\ \cline{1-3}
1k & 57.47 $\pm$ 1.78 & \textbf{61.37 $\pm$ 0.88} \\ \cline{1-3}
10k & 66.66 $\pm$ 0.88 & \textbf{68.06 $\pm$ 0.39} \\ \cline{1-3}
100k & 72.92 $\pm$ 0.21 &\textbf{73.19 $\pm$ 0.15}  \\ \cline{1-3}
\end{tabular}
\label{table:higgs_ssl}
\vspace{0cm}
\end{table}

Table \ref{table:higgs_ssl} shows that unsupervised pre-training significantly improves performance on the supervised classification task, especially in the regime where the unlabeled dataset is much larger than the labeled dataset. 
As exemplified in Fig. \ref{fig:ssl_training} the model convergence is much faster with unsupervised pre-training. Very fast convergence can be useful for continual learning and domain adaptation.

\vspace{0cm}
\section{Conclusions}
\vspace{0cm}
We have proposed TabNet, a novel deep learning architecture for tabular learning. 
TabNet uses a sequential attention mechanism to choose a subset of semantically meaningful features to process at each decision step. 
Instance-wise feature selection enables efficient learning as the model capacity is fully used for the most salient features, and also yields more interpretable decision making via visualization of selection masks. 
We demonstrate that TabNet outperforms previous work across tabular datasets from different domains. 
Lastly, we demonstrate significant benefits of unsupervised pre-training for fast adaptation and improved performance.

\section{Acknowledgements}

Discussions with Jinsung Yoon, Kihyuk Sohn, Long T. Le, Ariel Kleiner, Zizhao Zhang, Andrei Kouznetsov, Chen Xing, Ryan Takasugi and Andrew Moore are gratefully acknowledged.

\appendix

\section{Performance on KDD datasets}

\begin{table}[!htbp]
\caption{Performance on KDD datasets.}
\centering
\begin{tabular}{|C{1.3 cm}|C{1.3 cm}|C{1.3 cm}|C{1.3 cm}|C{1.3 cm}|}
\cline{1-5}
    \multirow{2}{*}{\textit{Model}}& \multicolumn{4}{|c|}{\textit{Test  accuracy (\%)}} \\  \cline{2-5}
    & \textit{Appetency} & \textit{Churn} & \textit{Upselling} & \textit{Census} \\ \cline{1-5}
    XGBoost & \textbf{98.2} & 92.7 & \textbf{95.1}  &\textbf{95.8}    \\ \cline{1-5}
    CatBoost  & \textbf{98.2} & \textbf{92.8} & \textbf{95.1}   & 95.7    \\  \hhline{|=|=|=|=|=|}
    \textit{TabNet}  &  \textbf{98.2}   &  92.7  &  95.0  &  95.5 \\ \cline{1-5}
\end{tabular}
\label{table:kdd_datasets}
\end{table}

Appetency, Churn and Upselling datasets are classification tasks for customer relationship management, and KDD Census Income \citep{UCI} dataset is for income prediction from demographic and employment related variables. 
These datasets show saturated behavior in performance (even simple models yield similar results). 
Table \ref{table:kdd_datasets} shows that TabNet achieves very similar or slightly worse performance than XGBoost and CatBoost, that are known to be robust as they contain high amount of ensembles.

\section{Comparison of feature importance ranking of TabNet}

\begin{table}[!htbp]
\caption{Importance ranking of features for Adult Census Income. TabNet yields feature importance rankings consistent with the well-known methods.}
\centering
\begin{tabular}{|C{2.2 cm}|C{0.9 cm}|C{0.9 cm}|C{1.1 cm}|C{0.9 cm}|}
\cline{1-5}
\emph{Feature} & \emph{SHAP} & \emph{Skater} & \emph{XGBoost} & \emph{TabNet} \\ \cline{1-5}
{Age} & 1 & 1 & 1 & 1\\ \cline{1-5}
{Capital gain} & 3 & 3 & 4 & 6\\ \cline{1-5}
{Capital loss} & 9 & 9 & 6 & 4\\ \cline{1-5}
{Education} & 5 & 2 & 3 & 2\\ \cline{1-5}
{Gender} & 8 & 10 & 12 & 8\\ \cline{1-5}
{Hours per week} & 7 & 7 & 2 & 7\\ \cline{1-5}
{Marital status} & 2 & 8 & 10 & 9\\ \cline{1-5}
{Native country} & 11 & 11 & 9 & 12\\ \cline{1-5}
{Occupation} & 6 & 5 & 5 & 3\\ \cline{1-5}
{Race} & 12 & 12 & 11 & 11\\ \cline{1-5}
{Relationship} & 4 & 4 & 8 & 5\\ \cline{1-5}
{Work class} & 10 & 8 & 7 & 10\\ \cline{1-5}
\end{tabular}
\label{table:importance_ranking}
\end{table}

We observe the commonality of the most important features (``Age", ``Capital gain/loss", ``Education number", ``Relationship") and the least important features (``Native country", ``Race", ``Gender", ``Work class"). 

\section{Self-supervised learning on Forest Cover Type}

\begin{table}[!htbp]
\caption{Self-supervised tabular learning results. Mean and std. of accuracy (over 15 runs) on Forest Cover Type, varying the size of the training dataset for supervised fine-tuning.}
\vspace{-0.3cm}
\centering
\begin{tabular}{|C{1.8 cm}|C{2.5 cm}|C{2.5 cm}|}
\cline{1-3}
\textit{Training} & \multicolumn{2}{|c|}{\textit{Test accuracy} (\%)}      \\ \cline{2-3}
\textit{dataset size}  & \textit{Supervised} & \textit{With pre-training} \\ \cline{1-3}
1k & 65.91 $\pm$ 1.02 & \textbf{67.86 $\pm$ 0.63} \\ \cline{1-3}
10k & 78.85 $\pm$ 1.24 & \textbf{79.22 $\pm$ 0.78} \\ \cline{1-3}
\end{tabular}
\label{table:covertype_ssl}
\end{table}

\section{Experiment hyperparameters}
\label{hyperparameters}

For all datasets, we use a pre-defined hyperparameter search space. $N_d$ and $N_a$ are chosen from $\{8, 16, 24, 32, 64, 128\}$, $N_{steps}$ is chosen from $\{3, 4, 5, 6, 7, 8, 9, 10\}$, $\gamma$ is chosen from $\{1.0, 1.2, 1.5, 2.0\}$, $\lambda_{sparse}$ is chosen from $\{0, 0.000001, 0.0001, 0.001, 0.01, 0.1\}$, $B$ is chosen from $\{256, 512, 1024, 2048, 4096, 8192, 16384, 32768\}$, $B_V$ is chosen from $\{256, 512, 1024, 2048, 4096\}$, the learning rate is chosen from $\{0.005,0.01.0.02,0.025\}$, the decay rate is chosen  from $\{0.4,0.8,0.9,0.95\}$ and the decay iterations is chosen from $\{0.5k,2k,8k,10k,20k\}$, and $m_B$ is chosen from $\{0.6, 0.7, 0.8, 0.9, 0.95, 0.98\}$. If the model size is not under the desired cutoff, we decrease the value to satisfy the size constraint. For all the comparison models, we run a hyperparameter tuning with the same number of search steps. 
\noindent\newline{\bf Synthetic:}
All TabNet models use $N_d{=}N_a{=}16$, $B{=}3000$, $B_V{=}100$, $m_B{=}0.7$. For Syn1 we use $\lambda_{sparse}{=}0.02$, $N_{steps}{=}4$ and $\gamma{=}2.0$; for Syn2 and Syn3 we use $\lambda_{sparse}{=}0.01$, $N_{steps}{=}4$ and $\gamma{=}2.0$; and for Syn4, Syn5 and Syn6 we use $\lambda_{sparse}{=}0.005$, $N_{steps}{=}5$ and $\gamma{=}1.5$. Feature transformers use two shared and two decision step-dependent FC layer, ghost BN and GLU blocks. All models use Adam with a learning rate of 0.02 (decayed 0.7 every 200 iterations with an exponential decay) for 4k iterations.
For visualizations, we also train TabNet models with datasets of size 10M samples. For this case, we choose $N_d=N_a=32$, $\lambda_{sparse}{=}0.001$, $B{=}10000$, $B_V{=}100$, $m_B{=}0.9$. Adam is used with a learning rate of 0.02 (decayed 0.9 every 2k iterations with an exponential decay) for 15k iterations. For Syn2 and Syn3, $N_{steps}{=}4$ and $\gamma{=}2$. For Syn4 and Syn6, $N_{steps}{=}5$ and $\gamma{=}1.5$. 
\noindent\newline{\bf Forest Cover Type:}
The dataset partition details, and the hyperparameters of XGBoost, LigthGBM, and CatBoost are from \citep{xgboost_gpu}. We re-optimize AutoInt hyperparameters.
TabNet model uses $N_d{=}N_a{=}64$, $\lambda_{sparse}{=}0.0001$, $B{=}16384$, $B_V{=}512$, $m_B{=}0.7$, $N_{steps}{=}5$ and $\gamma{=}1.5$. Feature transformers use two shared and two decision step-dependent FC layer, ghost BN and GLU blocks. Adam is used with a learning rate of 0.02 (decayed 0.95 every 0.5k iterations with an exponential decay) for 130k iterations. 
For unsupervised pre-training, the decoder model uses $N_d{=}N_a{=}64$, $B{=}16384$, $B_V{=}512$, $m_B{=}0.7$, and $N_{steps}{=}10$. For supervised fine-tuning, we use the batch size $B{=}B_V$ as the training datasets are small.
\noindent\newline{\bf Poker Hands:}
We split 6k samples for validation from the training dataset, and after optimization of the hyperparameters, we retrain with the entire training dataset. DT, MLP and deep neural DT models follow the same hyperparameters with \citep{dndt}. We tune the hyperparameters of XGBoost, LigthGBM, and CatBoost.
TabNet uses $N_d{=}N_a{=}16$, $\lambda_{sparse}{=}0.000001$, $B{=}4096$, $B_V{=}1024$, $m_B=0.95$, $N_{steps}{=}4$ and $\gamma{=}1.5$. Feature transformers use two shared and two decision step-dependent FC layer, ghost BN and GLU blocks. Adam is used with a learning rate of 0.01 (decayed 0.95 every 500 iterations with an exponential decay) for 50k iterations.
\noindent\newline{\bf Sarcos:}
We split 4.5k samples for validation from the training dataset, and after optimization of the hyperparameters, we retrain with the entire training dataset. All comparison models follow the hyperparameters from \citep{ant}.
TabNet-S model uses $N_d{=}N_a{=}8$, $\lambda_{sparse}{=}0.0001$, $B{=}4096$, $B_V{=}256$, $m_B{=}0.9$, $N_{steps}{=}3$ and $\gamma{=}1.2$. Each feature transformer block uses one shared and two decision step-dependent FC layer, ghost BN and GLU blocks. Adam is used with a learning rate of 0.01 (decayed 0.95 every 8k iterations with an exponential decay) for 600k iterations.
TabNet-M model uses $N_d{=}N_a{=}64$, $\lambda_{sparse}{=}0.0001$, $B{=}4096$, $B_V{=}128$, $m_B{=}0.8$, $N_{steps}{=}7$ and $\gamma{=}1.5$. Feature transformers use two shared and two decision step-dependent FC layer, ghost BN and GLU blocks. Adam is used with a learning rate of 0.01 (decayed 0.95 every 8k iterations with an exponential decay) for 600k iterations.
The TabNet-L model uses $N_d{=}N_a{=}128$, $\lambda_{sparse}{=}0.0001$, $B{=}4096$, $B_V{=}128$, $m_B{=}0.8$, $N_{steps}{=}5$ and $\gamma{=}1.5$. Feature transformers use two shared and two decision step-dependent FC layer, ghost BN and GLU blocks. Adam is used with a learning rate of 0.02 (decayed 0.9 every 8k iterations with an exponential decay) for 600k iterations.
\noindent\newline{\bf Higgs:}
We split 500k samples for validation from the training dataset, and after optimization of the hyperparameters, we retrain with the entire training dataset. MLP models are from \citep{sparse_mlp}. For gradient boosted trees \citep{tf_gbdt}, we tune the learning rate and depth -- the gradient boosted tree-S, -M, and -L models use 50, 300 and 3000 trees respectively.
TabNet-S model uses $N_d{=}24$, $N_a{=}26$, $\lambda_{sparse}{=}0.000001$, $B{=}16384$, $B_V{=}512$, $m_B{=}0.6$, $N_{steps}{=}5$ and $\gamma{=}1.5$. Feature transformers use two shared and two decision step-dependent FC layer, ghost BN and GLU blocks. Adam is used with a learning rate of 0.02 (decayed 0.9 every 20k iterations with an exponential decay) for 870k iterations.
TabNet-M model uses $N_d{=}96$, $N_a{=}32$, $\lambda_{sparse}{=}0.000001$, $B{=}8192$, $B_V{=}256$, $m_B{=}0.9$, $N_{steps}{=}8$ and $\gamma{=}2.0$. Feature transformers use two shared and two decision step-dependent FC layer, ghost BN and GLU blocks. Adam is used with a learning rate of 0.025 (decayed 0.9 every 10k iterations with an exponential decay) for 370k iterations.
For unsupervised pre-training, the decoder model uses $N_d{=}N_a{=}128$, $B{=}8192$, $B_V{=}256$, $m_B{=}0.9$, and $N_{steps}{=}20$. For supervised fine-tuning, we use the batch size $B{=}B_V$ as the training datasets are small.
\noindent\newline{\bf Rossmann:}
We use the same preprocessing and data split with \citep{catboost_rossmann} -- data from 2014 is used for training and validation, whereas 2015 is used for testing. We split 100k samples for validation from the training dataset, and after optimization of the hyperparameters, we retrain with the entire training dataset. The performance of the comparison models are from \citep{catboost_rossmann}. Obtained with hyperparameter tuning, the MLP is composed of 5 layers of FC (with a hidden unit size of 128), followed by BN and ReLU nonlinearity, trained with a batch size of 512 and a learning rate of 0.001.
TabNet model uses $N_d{=}N_a{=}32$, $\lambda_{sparse}{=}0.001$, $B{=}4096$, $B_V{=}512$, $m_B{=}0.8$, $N_{steps}{=}5$ and $\gamma{=}1.2$. Feature transformers use two shared and two decision step-dependent FC layer, ghost BN and GLU blocks. Adam is used with a learning rate of 0.002 (decayed 0.95 every 2000 iterations with an exponential decay) for 15k iterations.
\noindent\newline{\bf KDD:}
For Appetency, Churn and Upselling datasets, we apply the similar preprocessing and split as \citep{catboost}. The performance of the comparison models are from \citep{catboost}. TabNet models use $N_d{=}N_a{=}32$, $\lambda_{sparse}{=}0.001$, $B{=}8192$, $B_V{=}256$, $m_B{=}0.9$, $N_{steps}{=}7$ and $\gamma{=}1.2$. Each feature transformer block uses two shared and two decision step-dependent FC layer, ghost BN and GLU blocks. Adam is used with a learning rate of 0.01 (decayed 0.9 every 1000 iterations with an exponential decay) for 10k iterations.
For Census Income, the dataset and comparison model specifications follow \citep{online_bagging}. TabNet model uses $N_d{=}N_a{=}48$, $\lambda_{sparse}{=}0.001$, $B{=}8192$, $B_V{=}256$, $m_B{=}0.9$, $N_{steps}{=}5$ and $\gamma{=}1.5$. Feature transformers use two shared and two decision step-dependent FC layer, ghost BN and GLU blocks. Adam is used with a learning rate of 0.02 (decayed 0.7 every 2000 iterations with an exponential decay) for 4k iterations.
\noindent\newline{\bf Mushroom edibility:}
TabNet model uses $N_d{=}N_a{=}8$, $\lambda_{sparse}{=}0.001$, $B{=}2048$, $B_V{=}128$, $m_B{=}0.9$, $N_{steps}{=}3$ and $\gamma{=}1.5$. Feature transformers use two shared and two decision step-dependent FC layer, ghost BN and GLU blocks. Adam is used with a learning rate of 0.01 (decayed 0.8 every 400 iterations with an exponential decay) for 10k iterations.
\noindent\newline{\bf Adult Census Income:}
TabNet model uses $N_d{=}N_a{=}16$, $\lambda_{sparse}=0.0001$, $B{=}4096$, $B_V{=}128$, $m_B{=}0.98$, $N_{steps}{=}5$ and $\gamma{=}1.5$. Feature transformers use two shared and two decision step-dependent layer, ghost BN and GLU blocks. Adam is used with a learning rate of 0.02 (decayed 0.4 every 2.5k iterations with an exponential decay) for 7.7k iterations. 85.7\% test accuracy is achieved.

\section{Ablation studies}

\begin{table*}[!h]
\caption{Ablation studies for the TabNet encoder model for the forest cover type dataset.}
\centering
\begin{tabular}{|M{10 cm}|M{1.7 cm}|M{0.8 cm}|}
\cline{1-3}
    \textit{Ablation cases} & \textit{Test accuracy \%} (difference) & \textit{Model size}       \\ \cline{1-3}
    Base ($N_d=N_a=64$, $\gamma=1.5$, $N_{steps}=5$, $\lambda_{sparse}=0.0001$, feature transformer block composed of two shared and two decision step-dependent layers, $B=16384$)  &  96.99   &  470k \\ \cline{1-3}
    Decreasing capacity via number of units (with $N_d=N_a=32$)
    &  94.99 (-2.00) &  129k \\ \cline{1-3} 
    Decreasing capacity via number of decision steps (with $N_{steps}=3$)  
    &  96.22 (-0.77)  &  328k \\ \cline{1-3} 
    Increasing capacity via number of decision steps (with $N_{steps}=9$)  & 95.48 (-1.51)  &  755k \\ \cline{1-3} 
    Decreasing capacity via all-shared feature transformer blocks 
    &  96.74 (-0.25)  &  143k \\ \cline{1-3} 
    Increasing capacity via decision step-dependent feature transformer blocks 
    &  96.76 (-0.23) &  703k \\ \cline{1-3} 
    Feature transformer block as a single shared layer
    &  95.32 (-1.67) &  35k \\ \cline{1-3} 
    Feature transformer block as a single shared layer, with ReLU instead of GLU
    &  93.92 (-3.07) &  27k \\ \cline{1-3} 
    Feature transformer block as two shared layers
    &  96.34 (-0.66) &  71k \\ \cline{1-3} 
    Feature transformer block as two shared layers and 1 decision step-dependent layer
    &  96.54 (-0.45) &  271k \\ \cline{1-3} 
    Feature transformer block as a single decision-step dependent layer
    & 94.71 (-0.28) &  105k \\ \cline{1-3} 
    Feature transformer block as a single decision-step dependent layer, with $N_d{=}N_a{=}128$ 
    & 96.24 (-0.75) &  208k \\ \cline{1-3} 
    Feature transformer block as a single decision-step dependent layer, with $N_d{=}N_a{=}128$ and replacing GLU with ReLU
    & 95.67 (-1.32) &  139k \\ \cline{1-3} 
    Feature transformer block as a single decision-step dependent layer, with $N_d{=}N_a{=}256$ and replacing GLU with ReLU
    & 96.41 (-0.58) &  278k \\ \cline{1-3} 
    Reducing the impact of prior scale (with $\gamma=3.0$) 
    &  96.49 (-0.50) &  470k \\ \cline{1-3} 
    Increasing the impact of prior scale (with $\gamma=1.0$) 
    &  96.67 (-0.32) &  470k \\ \cline{1-3} 
    No sparsity regularization (with $\lambda_{sparse}=0$)  
    &  96.50 (-0.49) &  470k \\ \cline{1-3} 
    High sparsity regularization (with $\lambda_{sparse}=0.01$)
    &  93.87 (-3.12) &  470k \\ \cline{1-3} 
    Small batch size ($B=4096$)  &  96.42 (-0.57)  &  470k \\ \cline{1-3} 
\end{tabular}
\label{table:ablation}
\end{table*}

Table \ref{table:ablation} shows the impact of ablation cases. For all cases, the number of iterations is optimized on the validation set.

Obtaining high performance necessitates appropriately-adjusted model capacity based on the characteristics of the dataset. Decreasing the number of units $N_d$, $N_a$ or the number of decision steps $N_{steps}$ are efficient ways of gradually decreasing the capacity without significant degradation in performance. On the other hand, increasing these parameters beyond some value causes optimization issues and do not yield performance benefits. 
Replacing the feature transformer block with a simpler alternative, such as a single shared layer, can still give strong performance while yielding a very compact model architecture. This shows the importance of the inductive bias introduced with feature selection and sequential attention. 
To push the performance, increasing the depth of the feature transformer is an effective approach. While increasing the depth, parameter sharing between feature transformer blocks across decision steps is an efficient way to decrease model size without degradation in performance. We indeed observe the benefit of partial parameter sharing, compared to fully decision step-dependent blocks or fully shared blocks. We also observe the empirical benefit of GLU, compared to conventional nonlinearities like ReLU. 

The strength of sparse feature selection depends on the two parameters we introduce: $\gamma$ and $\lambda_{sparse}$. We show that optimal choice of these two is important for performance. A $\gamma$ close to 1, or a high $\lambda_{sparse}$ may yield too tight constraints on the strength of sparsity and may hurt performance. On the other hand, there is still the benefit of a sufficient low $\gamma$ and sufficiently high $\lambda_{sparse}$, to aid learning of the model via a favorable inductive bias. 

Lastly, given the fixed model architecture, we show the benefit of large-batch training, enabled by ghost BN \citep{ghost_batch_norm}. The optimal batch size for TabNet seems considerably higher than the conventional batch sizes used for other data types, such as images or speech. 

\vspace{-0.3cm}
\section{Guidelines for hyperparameters}
\vspace{-0.1cm}

We consider datasets ranging from $\sim$10K to $\sim$10M samples, with varying degrees of fitting difficulty. TabNet obtains high performance on all with a few general principles on hyperparameters:
\vspace{-0.3cm}
\begin{itemize}[leftmargin=*]
\item For most datasets, $N_{steps} \in [3, 10]$ is optimal. Typically, when there are more information-bearing features, the optimal value of $N_{steps}$ tends to be higher. On the other hand, increasing it beyond some value may adversely affect training dynamics as some paths in the network becomes deeper and there are more potentially-problematic ill-conditioned matrices. A very high value of $N_{steps}$ may suffer from overfitting and yield poor generalization.
\item Adjustment of $N_d$ and $N_a$ is an efficient way of obtaining a trade-off between performance and complexity. $N_d = N_a$ is a reasonable choice for most datasets. A very high value of $N_d$ and $N_a$ may suffer from overfitting and yield poor generalization.
\item An optimal choice of $\gamma$ can have a major role on the performance. Typically a larger $N_{steps}$ value favors for a larger $\gamma$.
\item A large batch size is beneficial -- if the memory constraints permit, as large as 1-10 \% of the total training dataset size can help performance. The virtual batch size is typically much smaller.
\item Initially large learning rate is important, which should be gradually decayed until convergence.
\end{itemize}



\bibliography{references.bib}

\end{document}